# Discovering Cyclic Causal Models with Latent Variables: A General SAT-Based Procedure

**Antti Hyttinen**
HIIT & Dept. of Computer Science
University of Helsinki
Finland

**Patrik O. Hoyer**
HIIT & Dept. of Computer Science
University of Helsinki
Finland

**Frederick Eberhardt**
Philosophy
California Institute of Technology
Pasadena, CA, USA

**Matti Järvisalo**
HIIT & Dept. of Computer Science
University of Helsinki
Finland

## Abstract

We present a very general approach to learning the structure of causal models based on d-separation constraints, obtained from any given set of overlapping passive observational or experimental data sets. The procedure allows for both directed cycles (feedback loops) and the presence of latent variables. Our approach is based on a logical representation of causal pathways, which permits the integration of quite general background knowledge, and inference is performed using a Boolean satisfiability (SAT) solver. The procedure is complete in that it exhausts the available information on whether any given edge can be determined to be present or absent, and returns "unknown" otherwise. Many existing constraint-based causal discovery algorithms can be seen as special cases, tailored to circumstances in which one or more restricting assumptions apply. Simulations illustrate the effect of these assumptions on discovery and how the present algorithm scales.

## 1 INTRODUCTION

One of the main goals in many fields of science is to identify the causal relations existing among some set of variables of interest. Such causal knowledge may be inferred both from experimental data (randomized controlled trials) and passive observational measurements. In general the information available from multiple such studies may need to be combined to obtain an accurate picture of the underlying system. In recent years, many approaches to this *causal discovery* problem have been suggested (Spirtes et al., 1999; Richardson and Spirtes, 1999; Schmidt and Murphy, 2009; Claassen and Heskes, 2010; Peters et al., 2010; Triantafillou et al., 2010), building on the framework of causal Bayes networks (Spirtes et al., 1993; Pearl, 2000). In this framework, causal relations among a set of variables $\mathbf{V}$ are represented by a directed graph $\mathcal{G}$ in which each variable is represented by a node in the graph, and an arrow from node $x$ to node $y$ indicates that $x$ is a direct cause of $y$ (with respect to $\mathbf{V}$).

Although causal models based on directed graphs are often restricted to be *acyclic*, causal feedback can be represented by permitting directed cycles in $\mathcal{G}$, i.e. directed paths from a node back to itself. In addition, unmeasured common causes of two or more nodes in $\mathbf{V}$ are commonly represented by allowing bi-directed arrows ($\leftrightarrow$) between any pair of confounded nodes.[1] (If there are no such confounders, the set $\mathbf{V}$ is said to be *causally sufficient*.) Thus, in the most general case of cyclic causal structures with latent variables, any pair of nodes $x, y \in \mathbf{V}$, with $x \neq y$, can be connected by any combination of the edges $x \rightarrow y$, $y \rightarrow x$, and $x \leftrightarrow y$ (see Figure 1 for examples).

One of the key theoretical concepts in causal models based on directed graphs is the notion of *d-separation*, due to Geiger et al. (1990). This is a graphical separation criterion that provides the structural counterpart to (conditional) independencies in the probability distribution generated by the model. D-separation is based on paths in the graph. Since a single pair of nodes can be connected by multiple edges, in our model space a *path* is defined as a sequence of consecutive edges in the graph, without any restrictions on the types or orientations of the edges involved.

**Definition 1 (D-separation)** *A path $p$ is said to be d-separated (or blocked) by a set of nodes $\mathbf{C}$ if and only if (i) $p$ contains a chain $i \rightarrow m \rightarrow j$ or a fork $i \leftarrow m \rightarrow j$ such that the middle node $m$ is in $\mathbf{C}$, or (ii) $p$ contains an inverted fork (or collider) $i \rightarrow m \leftarrow j$*

---

[1] In this representation a latent variable affecting more than two observed variables is represented by two-way confounders (bi-directed edges) between all pairs of nodes corresponding to the affected observed variables.

*such that the middle node m is not in* **C** *and such that no descendant of m is in* **C**. *A set* **C** *is said to d-separate x from y if and only if* **C** *blocks every path from x to y. (Pearl, 2000)*

When applying Definition 1 to graphs with bi-directed edges such as in Figure 1(b), the bidirected edge $z \leftrightarrow w$ can be viewed as a latent structure $z \leftarrow l_{zw} \rightarrow w$.

In acyclic models, such as causal Bayes networks, if two nodes $x$ and $y$ are d-separated given a set **C** then the corresponding random variables are statistically independent when conditioning on **C** in the probability distribution generated by the model. If there are no statistical independencies in the distribution other than those implied by d-separation applied to the underlying graph, the distribution is said to be *faithful* to the graph. Thus, under an assumption of faithfulness causal discovery procedures can use the outcomes of statistical independence tests, applied to the observed data, to infer d-separation and hence structural properties of the underlying graph. For example, if in a set of four variables $\mathbf{V} = \{x, y, z, w\}$ it is found that (i) $x$ is unconditionally independent of $y$, (ii) $x$ is independent of $w$ given $z$, (iii) $y$ is independent of $w$ given $z$, and (iv) no other unconditional independencies are found, then the well-known PC-algorithm (Spirtes et al., 1993) will infer that the underlying causal structure is the one in Figure 1(a).

While the correspondence between probabilistic independence and d-separation is known to hold generally for acyclic models (even when there are latent variables), the case is not as clear for cyclic models. The correspondence is known to hold for linear causal relations with Gaussian error terms, i.e. non-recursive linear Gaussian structural equation models (Spirtes, 1995), and can be extended to models with correlated error terms, which is one way to account for causally insufficient sets of variables. A general characterization of the parameterizations of cyclic models (with latent variables), for which the correspondence between d-separation and probabilistic independence holds, is not known (Pearl and Dechter, 1996; Neal, 2000).

Following the standard approach of non-parametric causal discovery algorithms, we use d-separation relations as the basic input to our procedure, but acknowledge that in the cyclic case only the linear Gaussian models are known to provide the appropriate correspondence with statistical independence. We allow for a set-up similar to the overlapping data sets approach of the ION-procedure (Tillman et al., 2009) in that we do not restrict ourselves to a single data set measured over some set of observed nodes, but can handle d-separation relations that were obtained from different (overlapping) sets $\mathbf{V}_i$ of nodes. Analogously to

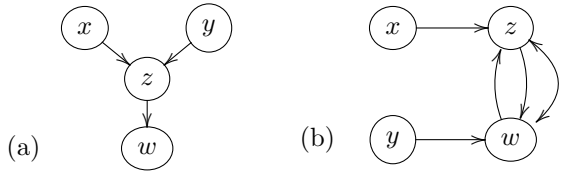

Figure 1: Example graphs (see text for details).

Hyttinen et al. (2012) we generalize the overlapping data sets case to allow that the $\mathbf{V}_i$ can contain nodes that are known to have been subject to a randomized experiment. Nevertheless, the target of our discovery procedure is the underlying causal graph $\mathcal{G}$ over the set of nodes $\mathbf{V} = \bigcup_i \mathbf{V}_i$, implying that $\mathcal{G}$ may contain edges between nodes that are never measured together in the same data set.

## 2 PROBLEM SETTING

We consider the space of cyclic causal models $\mathcal{G}$ over the (jointly) causally *in*sufficient set of nodes $\mathbf{V} = \bigcup_i \mathbf{V}_i$, where each $\mathbf{V}_i$ specifies the nodes present in experiment $\mathcal{E}_i = (\mathbf{J}_i, \mathbf{U}_i)$. $\mathbf{J}_i$ and $\mathbf{U}_i$ form a partition of $\mathbf{V}_i$ such that the nodes in $\mathbf{J}_i$ are randomized simultaneously and independently and the nodes in $\mathbf{U}_i$ are passively observed ($\mathbf{J}_i$ can be empty to allow for passive observational settings). We use the following simplification of the d-separation criterion:

**Definition 2 (D-separation)** *A path is d-connecting with respect to a conditioning set* **C** *if every collider c on the path is in* **C** *and no other nodes on the path are in* **C**, *otherwise the path is d-separated (or "blocked").*

Definition 2 is equivalent to Definition 1 when an edge can be used multiple times in a path (Studený, 1998; Koster, 2002). For example, the sequence of edges $x \rightarrow z \rightarrow w \leftarrow z \leftarrow y$ in the graph in Figure 1(a) is d-connecting with respect to conditioning set $\mathbf{C} = \{w\}$. The extension of d-separation to experimental settings is straightforward: a d-connecting path may only contain a node $x \in \mathbf{J}$ if $x \notin \mathbf{C}$ and $x$ is a fork (common cause) on the path or the source of the path. We write $x \perp y \,|\, \mathbf{C} \,||\, \mathbf{J}$ (resp. $x \not\perp y \,|\, \mathbf{C} \,||\, \mathbf{J}$) to denote that $x$ is d-separated from (resp. d-connected to) $y$ given **C** in the experiment with intervention set **J**. We assume we have a d-separation oracle that returns the truth values of statements of the form '$x \perp y \,|\, \mathbf{C} \,||\, \mathbf{J}_i$' in the true graph $\mathcal{G}$, for any pair of distinct nodes $x, y$ and set of nodes **C** that occur together in some $\mathbf{V}_i$.

It is well known that even in the presence of randomized experiments the set of all d-separation relations over the set of nodes in general underdetermines the true causal structure even for much more restricted model spaces than we consider here. So the discov-ery task is to determine for each pair of nodes in $\mathbf{V}$ and for each edge type $(\leftarrow, \rightarrow, \leftrightarrow)$ whether the edge is present, absent or if its existence is unknown. In addition we determine possible indirect (ancestral) causal relations: for each ordered pair of nodes $(x, y)$, whether a directed path $x \rightarrow \cdots \rightarrow y$ exists, does not exist or if its existence is unknown.

## 3 SAT AND BACKBONES

Our algorithm for causal structure discovery is based on computing the so-called backbone of a given formula in propositional logic. We employ a Boolean satisfiability (SAT) solver (Biere et al., 2009) to determine the backbone, which can be directly interpreted as the solution to the structure discovery task. This section provides an overview on SAT and backbones.

Propositional formulas are built from Boolean variables by repeated application of the connectives $\neg$ (negation), $\vee$ (disjunction, logical OR), $\wedge$ (conjunction, logical AND), $\Rightarrow$ (implication) and $\Leftrightarrow$ (equivalence). Any propositional formula can be represented in *conjunctive normal form* (CNF) using a standard linear-size encoding (Tseitin, 1983). For a Boolean variable $X$, there are two literals, $X$ and $\neg X$. A *clause* is a disjunction of literals; a CNF formula is a conjunction of clauses. A truth assignment is a function $\tau$ from Boolean variables to $\{0, 1\}$. A clause $C$ is satisfied by $\tau$ if $\tau(X) = 1$ for some literal $X$ in $C$, or $\tau(X) = 0$ for some literal $\neg X$ in $C$. A CNF formula $F$ is satisfiable if there is an assignment that satisfies all clauses in $F$, and unsatisfiable otherwise. The NP-complete Boolean satisfiability (SAT) problem asks whether a given CNF formula $F$ is satisfiable.

Implementations of decision procedures for SAT, so-called SAT solvers, can in practice not only determine satisfiability of CNF formulas, but also produce a satisfying truth assignment for satisfiable formulas. The most efficient SAT solvers are based on the complete conflict-driven clause learning (CDCL) search algorithm (Marques-Silva and Sakallah, 1999; Moskewicz et al., 2001; Eén and Sörensson, 2004). Central to CDCL is the ability to derive lemmas (in terms of new CNF clauses) based on non-solutions detected during search, which makes the search performed by CDCL SAT solvers differ from standard depth-first backtracking search. In many cases, the state-of-the-art CDCL SAT solvers can solve SAT instances consisting of millions of clauses and variables (Järvisalo et al., 2012).

If a Boolean variable $X$ takes the same value in all satisfying truth assignments of a given CNF formula $F$, $X$ is called a *backbone variable* of $F$; the value $X$ is assigned to in all satisfying assignments is called the polarity of $X$. The set of backbone variables (or simply, the backbone) of a formula $F$ can be computed by a linear number of calls (in the number of variables in $F$) to a SAT solver: if exactly one of $F \wedge X$ and $F \wedge \neg X$ is satisfiable, then $X$ is in the backbone of $F$.

## 4 ENCODING D-SEPARATION

Figure 2 shows our propositional encoding for the d-connection property. The encoding allows to represent both d-separation and d-connection relations as constraints directly on the edges present or absent in the underlying causal graph. In essence, the encoding spells out Definition 2 (extended to experiments) by expressing the conditions for paths being blocked or unblocked.

In the encoding, Boolean variables $[x \rightarrow y]$ and $[x \leftrightarrow y]$ represent the underlying causal graph. For each pair of distinct nodes $x, y \in \mathbf{V}$, the Boolean variable $[x \rightarrow y]$ (variable $[x \leftrightarrow y]$, respectively) takes the value 1 if and only if the edge $x \rightarrow y$ (edge $x \leftrightarrow y$, respectively) is present in the graph.[2] The Boolean variable $[x \not\perp y \,|\, \mathbf{C} \,||\, \mathbf{J}]$ is 1 if and only if $x$ and $y$ are d-connected in the underlying graph when conditioning on $\mathbf{C}$ and intervening on $\mathbf{J}$. To encode the different types of d-connecting paths of length $l$ between pairs of nodes $x, y$ when conditioning on $\mathbf{C}$ and intervening on $\mathbf{J}$ (Eqs. 1–7), Boolean variables $[x \,{-}\!\overset{l}{\cdots}\!{>}\, y]_{\mathbf{C},\mathbf{J}}$, $[x \,{<}\!\overset{l}{\cdots}\!{>}\, y]_{\mathbf{C},\mathbf{J}}$, and $[x \,{-}\!\overset{l}{\cdots}\!{-}\, y]_{\mathbf{C},\mathbf{J}}$ are introduced, with the respective arrowheads and edge-tails as indicated. In general, d-connecting paths in a cyclic graph can have infinite length, length of a path being the number of its edges. However, as shown in Appendix B, only paths of a maximum length $l_{\max} = 2|\mathbf{V}| - 4$ need to be considered. These Boolean variables are hence defined for all paths (of the four types) of length $l = 1, \ldots, l_{\max}$ and for all pairs of nodes in $\mathbf{V}$.

The constraint requiring that a specific d-connection $x \not\perp y \,|\, \mathbf{C} \,||\, \mathbf{J}$ is present is constructed by taking the conjunction of the variable $[x \not\perp y \,|\, \mathbf{C} \,||\, \mathbf{J}]$ and Equations 1–7. Similarly, the constraint requiring that a specific d-separation $x \perp y \,|\, \mathbf{C} \,||\, \mathbf{J}$ is present is the conjunction of $\neg[x \not\perp y \,|\, \mathbf{C} \,||\, \mathbf{J}]$ and Equations 1–7.

From a causal perspective, for a d-connection $x \not\perp y \,|\, \mathbf{C} \,||\, \mathbf{J}$ the encoding splits the d-connecting paths into four groups (Eq. 1): (i) paths that start with an edge-tail at $x$ and end with an arrowhead at $y$, (ii) paths that start with an arrowhead at $x$ and end with an edge-tail at $y$, (iii) paths that start with an arrowhead at $x$ and end with an arrowhead at $y$, and (iv)

---

[2] We omit self-loops, i.e. edges from a node to itself, as they do not affect the d-connectedness of a graph.

Encoding of d-connection between nodes $x, y$ given conditioning set $\mathbf{C}$ and intervention set $\mathbf{J}$.

$$[x \not\perp y \,|\, \mathbf{C} \,||\, \mathbf{J}] \quad \Leftrightarrow \quad \bigvee_{l=1}^{l_{\max}} \left( [x \xrightarrow[\mathbf{C},\mathbf{J}]{l} y] \vee [y \xrightarrow[\mathbf{C},\mathbf{J}]{l} x] \vee [x \xleftrightarrow[\mathbf{C},\mathbf{J}]{l} y] \vee [x \xlongdash[\mathbf{C},\mathbf{J}]{l} y] \right) \quad (1)$$

Paths of length 1:

$$[x \xrightarrow[\mathbf{C},\mathbf{J}]{1} y] \quad \Leftrightarrow \quad \begin{cases} [x \to y] & \text{if } y \notin \mathbf{J} \\ 0 & \text{otherwise} \end{cases} \quad (2)$$

$$[x \xleftrightarrow[\mathbf{C},\mathbf{J}]{1} y] \quad \Leftrightarrow \quad \begin{cases} [x \leftrightarrow y] & \text{if } x \notin \mathbf{J} \text{ and } y \notin \mathbf{J} \\ 0 & \text{otherwise} \end{cases} \quad (3)$$

$$[x \xlongdash[\mathbf{C},\mathbf{J}]{1} y] \quad \Leftrightarrow \quad 0 \quad (4)$$

Paths of length $l = 2, \ldots, l_{\max}$:

$$[x \xrightarrow[\mathbf{C},\mathbf{J}]{l} y] \Leftrightarrow \bigvee_{z \notin \mathbf{C}} \left([x \xrightarrow[\mathbf{C},\mathbf{J}]{1} z] \wedge [z \xrightarrow[\mathbf{C},\mathbf{J}]{l-1} y]\right) \vee \bigvee_{z \in \mathbf{C}} \left([x \xrightarrow[\mathbf{C},\mathbf{J}]{1} z] \wedge [z \xleftrightarrow[\mathbf{C},\mathbf{J}]{l-1} y]\right) \quad (5)$$

$$[x \xleftrightarrow[\mathbf{C},\mathbf{J}]{l} y] \Leftrightarrow \bigvee_{z \notin \mathbf{C}} \left([z \xrightarrow[\mathbf{C},\mathbf{J}]{1} x] \wedge [z \xrightarrow[\mathbf{C},\mathbf{J}]{l-1} y]\right) \vee \bigvee_{z \notin \mathbf{C}} \left([z \xrightarrow[\mathbf{C},\mathbf{J}]{1} x] \wedge [z \xleftrightarrow[\mathbf{C},\mathbf{J}]{l-1} y]\right) \vee$$

$$\bigvee_{z \notin \mathbf{C}} \left([x \xleftrightarrow[\mathbf{C},\mathbf{J}]{1} z] \wedge [z \xrightarrow[\mathbf{C},\mathbf{J}]{l-1} y]\right) \vee \bigvee_{z \in \mathbf{C}} \left([x \xleftrightarrow[\mathbf{C},\mathbf{J}]{1} z] \wedge [z \xleftrightarrow[\mathbf{C},\mathbf{J}]{l-1} y]\right) \quad (6)$$

$$[x \xlongdash[\mathbf{C},\mathbf{J}]{l} y] \Leftrightarrow \bigvee_{z \notin \mathbf{C}} \left([x \xrightarrow[\mathbf{C},\mathbf{J}]{1} z] \wedge [z \xlongdash[\mathbf{C},\mathbf{J}]{l-1} y]\right) \vee \bigvee_{z \in \mathbf{C}} \left([x \xrightarrow[\mathbf{C},\mathbf{J}]{1} z] \wedge [y \xrightarrow[\mathbf{C},\mathbf{J}]{l-1} z]\right) \quad (7)$$

Figure 2: Encoding d-connection via paths between pairs of nodes.

paths that start with an edge-tail at $x$ and end with an edge-tail at $y$. The paths are built up recursively in terms of length $l$ (Eqs. 5, 6, and 7). By keeping track of the path lengths we ensure that each path bases out through Eqs. 2 and 3 on the actual edges in the graph, whose presence is represented by Boolean variables $[x \to y]$ and $[x \leftrightarrow y]$. There are no paths of type (iv) with length 1, as such a path must involve at least one collider (in $\mathbf{C}$) to have tails at both ends (hence Eq. 4). The shortest valid case is of length $l = 2$ and results from the second half of Eq. 7. By explicitly keeping track of the terminal edge-marks in each path variable, the encoding ensures that all colliders on a d-connecting path are in the conditioning set $\mathbf{C}$, and all non-colliders are not in $\mathbf{C}$. The base cases (Eqs. 2 and 3) ensure that there is no path with an edge into a variable that is intervened on (into $y \in \mathbf{J}$).

For each given d-separation relation $x \perp y \,|\, \mathbf{C} \,||\, \mathbf{J}$ (or similarly each d-connection relation $x \not\perp y \,|\, \mathbf{C} \,||\, \mathbf{J}$), the whole encoding, including Eqs. 1–7, is cubic in the number $|\mathbf{V}|$ of nodes. Furthermore, it is important to notice that our algorithm, as described next, does not generate the constraints in Eqs. 1–7 for all possible d-separation and d-connection relations at once. The constraints for individual relations are generated only on demand during the execution of the algorithm, in many cases avoiding generating an exponential number of constraints needed to represent all possible d-separation and d-connection relations.

The SAT-based approach to causal structure discovery by Triantafillou et al. (2010) uses an encoding based on partial ancestral graphs (PAGs), a particular form of equivalence class. Their encoding does not suffice for our purposes, since it is restricted to *acyclic* causal structures in *non-experimental* settings, and given experiments it is often possible to distinguish between different graphs that for passive observational data belong to the same PAG.

## 5 ALGORITHM

The encoding of d-separation relations presented in the previous section can be used for a variety of discovery applications. For the purpose of illustration we will present here one algorithm for a common discovery setting. The extension to other settings is then easily explained. Algorithm 1 iterates over three steps until all d-separation relations are known: (1) finding a set of d-separation/connection tests $T_c$ (in order of increasing conditioning set size) with currently unknown result, and determining those relations $D_c$, (2) generating the additional constraints encoding the relations in $D_c$ (recall the encoding in Figure 2), and (3) computing the backbone over the propositional formula

**Algorithm 1** SAT-based causal structure discovery
---

Initialize solution $S$ for the edge variables $[x \to y]$, $[x \leftrightarrow y]$ of each pair of distinct nodes $x, y \in \mathbf{V}$ to status *unknown*.

Initialize $\varphi$ to be the empty propositional formula.

For conditioning set size $c$ from 0 to $|\mathbf{V}| - 2$:

  1: *Determine d-separation/connection relations.*
  Find a set $T_c$ of d-separation/connection tests with conditioning set size $c$ that are undetermined given $S$.
  Determine the d-separation/connection relation for each test in $T_c$, and let set $D_c$ consist of these relations.

  2: *Refine the working formula $\varphi$.*
  For each $x \perp y \,|\mathbf{C}\,||\,\mathbf{J}$ in $D_c$:
    Encode $x \perp y \,|\mathbf{C}\,||\,\mathbf{J}$ *using equations 1-7*:
    $\varphi := \varphi \wedge$ Encode$(x \perp y \,|\mathbf{C}\,||\,\mathbf{J})$.
  For each $x \not\perp y \,|\mathbf{C}\,||\,\mathbf{J}$ in $D_c$:
    Encode $x \not\perp y \,|\mathbf{C}\,||\,\mathbf{J}$ *using equations 1-7*:
    $\varphi := \varphi \wedge$ Encode$(x \not\perp y \,|\mathbf{C}\,||\,\mathbf{J})$.

  3: *Incremental backbone computation with SAT solver*
  Compute $B$: the set of edge-variables $[x \to y]$, $[x \leftrightarrow y]$ in the backbone of $\varphi$.
  For each edge variable $e$ in $B$:
    If $e \in B$ with polarity 1, set status of $e$ to *present* in $S$.
    If $e \in B$ with polarity 0, set status of $e$ to *absent* in $S$.

Output $S$: the status of each edge.

---

consisting of the constraints generated so far.

In Step 1 we apply a pruning heuristic (described in Appendix A) that guarantees that all unknown d-separation relations are found, but remains computationally tractable. We use a d-separation oracle (see Section 2) to determine the result of each test.

In Step 2, given a d-connection relation $x \not\perp y \,|\mathbf{C}\,||\,\mathbf{J}$, the subroutine Encode returns the conjunction of $[x \not\perp y \,|\mathbf{C}\,||\,\mathbf{J}]$ and the formulas in Eqs. 1–7. Similarly, given a d-separation relation $x \perp y \,|\mathbf{C}\,||\,\mathbf{J}$, Encode returns the conjunction of $\neg[x \not\perp y \,|\mathbf{C}\,||\,\mathbf{J}]$ and the formulas in Eqs. 1–7. Note that for each combination of $\mathbf{C}$ and $\mathbf{J}$, Eqs. 2–7 need to the added only once into $\varphi$ (also guaranteed by our current implementation). This is important in practice, so that the SAT solver is not suffocated with many copies of the same constraints.

In Step 3, a SAT solver is used incrementally for determining which of the edge-variables in the current working formula $\varphi$ are in the backbone of $\varphi$. The polarity of these backbone variables determines whether the corresponding edges are present or absent.

Like other constraint based causal discovery algorithms, Algorithm 1 considers d-separation relations in order of the size of the conditioning set $\mathbf{C}$. For sparse graphs, this enables a rapid pruning of the constraint generation on the basis of the simplest tests.

But unlike other constraint based algorithms, Algorithm 1 can explicitly include known d-connections, rather than assuming that there is a d-connection whenever no d-separation is found (see also Section 7).

Algorithm 1 is easily adjusted to consider an *arbitrary set* of d-separation/connection relations as input, as long as the set of nodes $\mathbf{V}$ is specified from the outset. If the set is small, one can just run step 2 and 3 to compute the backbone using all available relations, otherwise one can run the full procedure, simply omitting relations from $D_c$ that are not available in the set. It will terminate when all relations are encoded or when no more are needed, as determined by step 1.

In Algorithm 1 we use the status on each edge as the output. If other aspects of the graphs are of interest, one can easily define other variables and compute the backbone over them. In Section 6 we use this feature to determine which ancestral relationships are known.

### 5.1 BACKGROUND KNOWLEDGE AND MODEL SPACE ASSUMPTIONS

Although we have considered a very general model space, restricting the procedure to smaller spaces is simple. Focusing on just one data set rather than a set of overlapping data sets, or only considering passive observational data and no experiments, requires no adjustments of Algorithm 1. If one has reason to believe that there are no unmeasured nodes, i.e. that $\mathbf{V}$ is (jointly) causally sufficient, then setting

$$[x \leftrightarrow y] \Leftrightarrow 0 \qquad (8)$$

for all pairs of nodes in the encoding will enforce this restriction. If one is only interested in acyclic causal structures, then adding the constraint

$$\neg[x \not\perp y \,|\emptyset\,||\,\{x\}] \vee \neg[x \not\perp y \,|\emptyset\,||\,\{y\}] \qquad (9)$$

for each pair of nodes, together with the respective path definitions (Eqs. 2–7), is sufficient. Eq. 9 disallows cycles by enforcing that there cannot be a directed path from $x$ to $y$ *and* a directed path from $y$ to $x$. Since the conditioning set in each of the d-connection claims in Eq. 9 is empty, there cannot exist any colliders in the d-connecting paths. The intervention on $x$ and $y$, respectively, in each of the claims in Eq. 9 ensures that d-connections due to common causes are excluded. Only *directed* paths are involved in $x \not\perp y \,|\emptyset\,||\,\{x\}$ and $x \not\perp y \,|\emptyset\,||\,\{y\}$. In Section 5.2 we use this flexibility to generate the same causal inferences as other d-separation based algorithms.

More generally, the encoding allows for including various types of background knowledge. One can enforce that a particular edge is present or absent, that particular ancestral relations are maintained or disallowed, that specific paths (with, if needed, particular waypoints and of a specific length) are present or absent. The type of knowledge that can be encoded is more general than any other constraint based procedure we are aware of, including the additions to the cSAT+ algorithm by Borboudakis et al. (2011). One is in principle only limited by what can be encoded in terms of a Boolean constraint over the edge and path variables. We think this could be of enormous utility to applications with significant domain knowledge or when qualitative causal relations are discovered by other means (e.g. using the additive noise or non-Gaussian techniques of Peters et al. (2010) and Hoyer et al. (2008)).

## 5.2 COMPLETENESS

For more restricted model spaces, graphical representations of the classes of d-separation-equivalent graphs have been developed (e.g. partial ancestral graphs). We do not have a similar representation for our more general model space and it is unclear whether an easily interpretable representation is possible, since there can be graphs that share the same d-separation relations, but differ in adjacencies, orientations *and* ancestral relationships. By only providing the status of each edge as output of Algorithm 1, we follow Triantafillou et al. (2010) who used this solution format in light of the often (even computationally) unmanageable output of the ION-algorithm (which does not consider cyclic graphs). The downside is that this output is not fully informative about the solution space. For example, if d-separation relations were obtained from a passive observation of the graph $x \to y \to z$, then the current output does not represent that $x \to y \leftarrow z$ is not a solution. Instead, it would (among other things) mark all edges of adjacent nodes as unknown, since $x \leftarrow y \leftarrow z$ is also a solution. Nevertheless, it is trivial to query our procedure about graphical aspects that are not represented in the output. Since the complete solution space is implicitly represented by the working formula $\varphi$, the SAT-solver can easily determine that $x \to y \leftarrow z$ is not a valid solution in this example. Similarly, one can query the status of any other structural proposition by constructing a Boolean variable $X$ for it using the edge or path variables in the encoding, and determining whether $X$ is in the backbone of $\varphi$ or not. If it is, then polarity 1 indicates that $X$ is true for *all* graphs that satisfy $\varphi$, while polarity 0 indicates that $X$ is false for *all* graphs that satisfy $\varphi$. If $X$ is not in the backbone, then there is a graph $\mathcal{G}_1$ that satisfies $\varphi$, for which $X$ is true, and a graph $\mathcal{G}_2$ that satisfies $\varphi$, but for which $X$ is false. This is one, given the encoding perhaps trivial, sense in which our procedure is complete for any propositional query given the d-separation/connection relations (and any model space restrictions) as input. We call this *query-completeness*.

A different type of completeness is used in the context of other constraint based algorithms. Given the d-separation tests that an algorithm performs, we say that an algorithm is *d-separation complete* if all d-separation relations over the set of nodes are known. The PC-algorithm (for acyclic graphs over a causally sufficient set of nodes), the FCI-algorithm (for acyclic graphs over a causally *in*sufficient set of nodes) and the CCD-algorithm (for *cyclic* graphs over a causally sufficient set of nodes) are all d-separation complete for their model spaces, respectively (Spirtes et al., 1993; Richardson, 1996; Spirtes et al., 1999). Relying on the model space assumptions, the algorithms conduct just enough d-separation tests to determine *all* d-separation relations of the graphs in the solution space, even relations that the algorithms did not explicitly test. None of these algorithms are d-separation complete when their model space assumptions are violated: Figure 1(b) gives a cyclic graph for which FCI is not d-separation complete, since it does not test whether $x \perp y \,|\, \{w, z\}$. The graph with latent confounders in Figure 3 is an example for which CCD is not d-separation complete, because it does not determine the d-separation $x_1 \perp x_5 \,|\, \{x_2, x_3, x_4\}$. PC does not handle either graph. These limitations illustrate that achieving d-separation completeness without performing all tests is a non-trivial problem in the general model space we consider (containing both graphs). Once we consider overlapping data sets, there are d-separation relations involving nodes that do not occur together in any $\mathbf{V}_i$. Sometimes these can be determined from the other d-separation relations, but often they remain undetermined even when all the d-separation relations within each $\mathbf{V}_i$ are established. For this setting we adjust the definition of d-separation completeness to require that exactly those relations that cannot be determined (in the sense just described) are left *unknown* and all others are determined.

For cyclic models with latent variables in overlapping experimental or observational data sets, Algorithm 1 *is* d-separation complete, and in general it will not test all available d-separation relations. But in the present implementation (of step 1) we resort to simple safe heuristics to avoid some redundant tests, and otherwise apply brute force (see Appendix A). It is an open challenge to further reduce the number of tests performed while preserving d-separation completeness. We cannot employ a simple variant of the efficient test schedules of FCI and CCD, as they select subsequent tests on the basis of a graphical representation of the knowledge acquired so far that is specific to their restricted model spaces. But *given* those restrictions, we can adopt the test schedules: Using FCI as basis, the ION-algorithm (which also assumes acyclicity) is d-separation complete for pas-sively observed overlapping data sets (Tillman et al., 2009). Similarly, if we assume acyclicity, we can use instead of our heuristic the test schedule of FCI in Algorithm 1 when analyzing overlapping *experimental* data sets: run FCI on each individual data set (experimental or not) and input to Algorithm 1 the results of the tests that FCI considered on each individual data set (together with the acyclicity constraints in (9) for the FCI model space). Algorithm 1 will then combine the information across data sets and output all information available on the status of each edge in the true graph. The set of d-separation relations tested by FCI is sufficient for d-separation completeness for the $\mathbf{V}_i$ in that data set. Interventions do not affect the d-separation completeness, since the manipulated graph in any experiment still satisfies all model space assumptions of FCI. (One could avoid some tests by further book-keeping of the information about the interventions, but for d-separation completeness it is unnecessary.) Given FCI's d-separation completeness on each data set, the constraints generated by feeding the test results to Algorithm 1 imply that all d-separations relations that could be tested, are already determined. Any d-separation relation still unknown cannot be determined. By assuming acyclicity we thus obtain d-separation completeness *using the efficient FCI schedule of tests* for overlapping data sets with experiments. As Algorithm 1 is also query-complete, we have a general procedure for the approaches of Lagani et al. (2012).

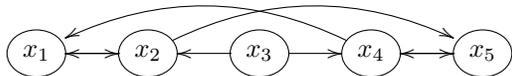

Figure 3: An (acyclic) graph with latents for which the CCD-algorithm is not d-separation complete.

An analogous argument for *cyclic* graphs *without* latent nodes, using the test schedule of CCD, can only be made if we assume that the nodes in $\mathbf{V}$ are all observed in all (possibly experimental) data sets. In the overlapping setting, causal sufficiency can be violated in the individual data sets and, as shown above, CCD is not d-separation complete for such a model space.

## 6 SIMULATIONS

To determine the effectiveness of the proposed approach, we implemented Algorithm 1 and investigated the properties of the method empirically. Our implementation is based on the MiniSAT solver (Eén and Sörensson, 2004, 2003). The code is available at http://www.cs.helsinki.fi/group/neuroinf/nonparam/.

First, we investigated the extent to which our approach, and in particular the SAT solver used, is able to solve the large problem instances generated by non-trivial graphs. We generated random directed graphs of size $n = 5 \ldots 12$ nodes, in which each of the edges (both directed and bidirected) was independently included with probability 0.2. We then generated a random set of 10 overlapping experiments, in each of which each node was independently and with equal probability chosen to be either intervened, passively observed, or unobserved. Finally, we computed all observable d-separation/connection relations; these constituted the input to our procedure.

Figure 4(a) gives, for each value of $n$, the median runtime based on 100 random problem instances, for the complete procedure (solid curve), and when only considering conditioning sets $\mathbf{C}$ with two or fewer elements (dotted curve). Note that most instances involving a relatively small number of nodes (on the order of 10 or less) can be solved by the complete procedure in minutes, if not seconds. We emphasize that these are not trivial problems: No other existing causal discovery procedure can handle our model space (allowing both latents and cycles), nor our very general experimental setup (overlapping data sets including interventions). At the same time, it is quite clear that, at least in its current implementation, the method does not scale to much larger numbers of variables. Scalability could be achieved with a more efficient search for unknown d-separations in Step 1 of the algorithm.

An effective way to reduce the run-time of the algorithm is to limit the size of the conditioning sets considered (dotted line in Figure 4(a)). While this means that completeness is not guaranteed, Figure 4(b) shows that in most cases very little is lost in terms of identifiability. We randomly sampled 100 problem instances as above, except that we now fixed the number of nodes to $n = 8$. The red solid curve shows the proportion of true directed edges (i.e. $x \rightarrow y$ in the true graph) which were identified as a direct edge (as opposed to unknown, since no errors are made). Similarly, the red dashed curve shows the identification of absences of direct edges, and the remaining curves indicate the amount of bidirected edges and existence of directed paths (ancestral relationships) identified. A key observation is that tests of higher order (roughly $|\mathbf{C}| \geq 3$) provide very little additional information over those involving smaller conditioning sets.

Finally, we investigated the extent to which our very general model space (allowing both cycles and latents) is detrimental to identification when the true model is more restricted. We generated a total of 300 random problem instances using the same procedure as above, each with $n = 8$ nodes, where the first 100 models were restricted to being acyclic, the second 100 were restricted to contain no latents (i.e. no edges of the form $x \leftrightarrow y$ in the true graph over $\mathbf{V}$), and the remaining

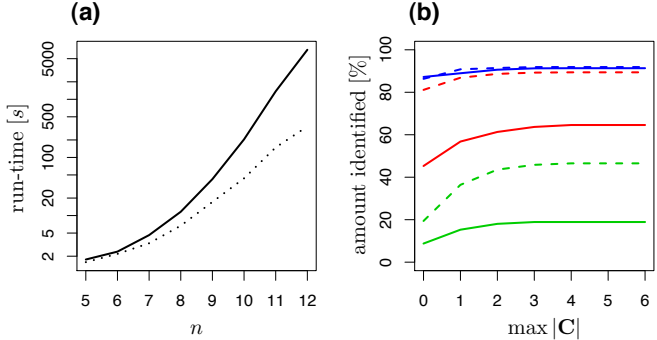
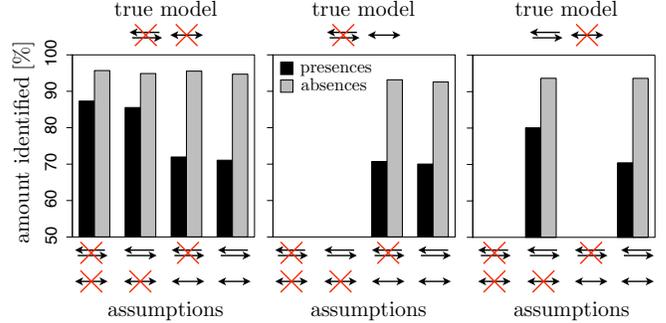

Figure 4: **(a)** Median run-time of the procedure as a function of the total number of nodes in the model. The dotted line gives the median run-time when restricting to $\max |\mathbf{C}| = 2$. **(b)** Proportion of edges (solid lines) and absences of edges (dashed lines) identified, as a function of $\max |\mathbf{C}|$. Directed edges are shown in red, bidirected edges (confounders) in green, and directed paths (ancestral relationships) in blue.

Figure 5: Proportion of directed edge presences and absences identified, under various model space assumptions, for acyclic true models without latents (left), acyclic models with latents (center), and cyclic models without latents (right).

100 were both acyclic and contained no latents. Figure 5 shows the proportion of direct edges identified, and the proportion of absences of direct edges identified, as a function of the assumptions used (assuming an acyclic model, assuming no latents, assuming both, or assuming neither). The general message is that very little identifiability seems to be lost when assuming the more general model spaces in this experimental setup.

## 7 DISCUSSION

By focusing exclusively on d-separation and d-connection relations obtained *without errors* we have so far taken the approach used by other constraint-based algorithms in the literature (PC, FCI, CCD, ION, IOD, cSAT+ etc.) to separate the causal from the statistical inference. As an important direction for future work, we now briefly discuss integrating statistical inference.

In most realistic situations d-separation/connection relations are determined by independence tests from statistical data. Such tests, especially when performed in large numbers, produce errors due to the finite number of samples available and problems of multiple testing. All other constraint-based causal discovery algorithms face similar problems. In our case, the errors can result in d-separation/connection relations that are contradictory. Since the logical encoding is simply unsatisfiable in such cases, no output is given. But there are more interesting features of the encoding and the algorithm that hold promise to be useful with actual statistical data. First, since no definite answer is required of a d-separation test, we can enforce different p-value thresholds to detect independencies and dependencies (see Tsamardinos et al. (2012)). If a p-value of a test falls between the thresholds, the d-separation relation can be treated as unknown, by just not adding any constraints into the working formula $\varphi$. This approach does not completely avoid conflicts, but reduces their number and allows for at least some more control than many extant algorithms are able to offer. A second approach to dealing with statistical issues would be to exploit extensions of SAT, especially Boolean optimization in terms of maximum satisfiability (MaxSAT) of propositional formulas (Biere et al., 2009), where the task is to find a truth assignment that satisfies the maximum number of CNF clauses. Hence a MaxSAT solver could be used for discovering causal models that entail a minimal number of contradictory d-separation/connection relations in the input.

## 8 CONCLUSION

We presented a causal discovery procedure for a very general model space: to our knowledge, it is currently the only nonparametric causal discovery algorithm that allows for a model space that includes graphs with cycles *and* latent confounders (recall the discussion on cycles and d-separation in Section 1). The algorithm can be applied to overlapping data sets, whether they are experimental or passive observational, and can incorporate a large variety of different background information if available. It does not depend on parametric restrictions such as linearity (Hyttinen et al., 2012), and requires only the ability to test for d-separation/connection relations.

SAT-based procedures have been previously proposed for the more restricted space of acyclic causal models (Triantafillou et al., 2010; Borboudakis and Tsamardinos, 2012). However, ours is the first procedure that is complete with respect to overlapping surgical experiments, and additionally handles a model space that allows for cycles. In order to capture the more general model space, we employ a novel logical encoding of d-separation and d-connection relations. The Boolean constraints for individual relations are generated iteratively and only on demand during the execution of our procedure, and an incremental SAT solver is used for iteratively computing the backbone of the Boolean constraints. Our procedure can also be easily used for the more restricted model spaces by introducing additional Boolean constraints. By constraining the model space to causally sufficient or acyclic causal structures we can perform the inferences of the standard algorithms in the literature, such as PC, FCI, ION, IOD, cSAT+ and CCD for moderately sized graphs. The inferences made are complete in the most general and in the more restricted settings.

## A  PRUNING HEURISTICS

In the (intermediary) solution $S$ describing our current knowledge some edges are present, some are absent and the presence of some edges is unknown. We consider two graphs $\mathcal{G}_1$ and $\mathcal{G}_2$, such that they agree on all the edges that are determined, but $\mathcal{G}_1$ omits all undetermined edges, while $\mathcal{G}_2$ includes all undetermined edges as present. As removing edges can only result in more d-separation relations, a d-connection relation present in $\mathcal{G}_1$ must be present in all solutions. Similarly, a d-separation relation present in $\mathcal{G}_2$ must be present in all possible solutions. Only the remaining tests are possibly informative. This is a safe heuristic that turns out to be computationally feasible, as forward calculation of d-separation/connection relations for a fully defined graph is fast for the model sizes we are considering. In addition, we also omit tests with conditioning sets that contain nodes that cannot be on a d-connecting path between the nodes in question.

## B  LIMIT ON THE PATH LENGTH

Written solely in terms of edge variables, the right-hand side of Eq. 1 is a large disjunction of d-connecting paths up to length $l_{\max}$ for the relation on the left-hand side. As a path of arbitrary length can be d-connecting, $l_{\max}$ should be infinite to guarantee soundness of the formulation. Here we show that only paths of lengths up to a certain upper bound need to be considered. The following lemma, proven at the end of this appendix, is essential in showing this.

**Lemma 1** *If there exists a path that is d-connecting with respect to $x \not\perp y \,|\, \mathbf{C}\,||\,\mathbf{J}$ and longer than $2|\mathbf{V}| - |\mathbf{C} \cup \mathbf{J} \cup \{x,y\}| - 1$ edges, then there exists a shorter path that is d-connecting with respect to the same relation.*

Consider a path $p_{\text{long}}$ that is d-connecting for $x \not\perp y \,|\, \mathbf{C}\,||\,\mathbf{J}$ and longer than $2|\mathbf{V}| - |\mathbf{C} \cup \mathbf{J} \cup \{x,y\}| - 1$. By Lemma 1 there is a path $p_{\text{short}}$ with at most length $2|\mathbf{V}| - |\mathbf{C} \cup \mathbf{J} \cup \{x,y\}| - 1$ edges that is d-connecting with respect to the same relation. Now the expanded version of the right hand side of Equation 1 has the form: $\ldots \vee [p_{\text{short}}] \vee [p_{\text{long}}] \vee \ldots$. The only situation where such a constraint may have a different value than $\ldots \vee [p_{\text{short}}] \vee \ldots$ is when $p_{\text{long}}$ exists and $p_{\text{short}}$ does not. This is impossible by the construction of $p_{\text{short}}$ using Lemma 1. We can thus ignore $[p_{\text{long}}]$ and all paths longer than $2|\mathbf{V}| - |\mathbf{C} \cup \mathbf{J} \cup \{x,y\}| - 1$. We can set $l_{\max} = 2|\mathbf{V}| - 4$, since if $\mathbf{C} = \emptyset$, then paths have at most length $|\mathbf{V}| - 1$.

**Proof of Lemma 1**  The following six rules always give a shorter d-connecting path with respect to the same relation. The deleted part is underlined on the left. Circles indicate arrowhead or tail.

$$\underline{x\!\circ\!\cdots\!\circ\!x}\!\circ\!\cdots\!\circ\!y \;\;\Rightarrow\;\; x\!\circ\!\cdots\!\circ\!y \tag{10}$$

$$x\!\circ\!\cdots\!\circ\!\underline{y\!\circ\!\cdots\!\circ\!y} \;\;\Rightarrow\;\; x\!\circ\!\cdots\!\circ\!y \tag{11}$$

$$\cdots\!>\!\underline{z\!<\!\cdots\!>}\!z\!<\!\cdots \;\;\Rightarrow\;\; \cdots\!>\!z\!<\!\cdots \tag{12}$$

$$\cdots\!>\!\underline{z\!-\!\cdots\!\circ}\!z\!-\!\cdots \;\;\Rightarrow\;\; \cdots\!>\!z\!-\!\cdots \tag{13}$$

$$\cdots\!-\!\underline{z\!\circ\!\cdots\!-}\!z\!<\!\cdots \;\;\Rightarrow\;\; \cdots\!-\!z\!<\!\cdots \tag{14}$$

$$\cdots\!-\!\underline{z\!\circ\!\cdots\!\circ}\!z\!-\!\cdots \;\;\Rightarrow\;\; \cdots\!-\!z\!-\!\cdots \tag{15}$$

The rules imply that if a middle node $z$ appears three times on a d-connecting path, the path will necessarily have at least one of the forms on the left in (12-15). (A path can never be d-connecting if the same node appears both as a collider and a non-collider somewhere on the path.) Thus a node can appear at most two times in paths that cannot be shortened.

First, consider the case with no colliders on the path. The only situation where a d-connecting path cannot be shortened and a node appears twice, occurs when the path has the form $\cdots\!>\!z\!-\!\cdots\!-\!z\!<\!\cdots$. This path cannot be d-connecting without a collider between the instances of $z$. Thus, without colliders a path that cannot be shortened has at most $|\mathbf{V}|$ nodes and thus length $|\mathbf{V}| - 1$. Second, if the path cannot be shortened, each node in $\mathbf{C} \cup \mathbf{J} \cup \{x,y\}$ can appear at most once due to (10-15). The remaining nodes can appear at most twice. This makes a total of $2|\mathbf{V}| - |\mathbf{C} \cup \mathbf{J} \cup \{x,y\}|$ nodes. Hence the length of the path is at most $2|\mathbf{V}| - |\mathbf{C} \cup \mathbf{J} \cup \{x,y\}| - 1$.  $\square$

### Acknowledgements

This research was supported by the Academy of Finland under grants 218147 and 255625 (POH), 132812 and 251170 (MJ), by HIIT (AH) and by the James S. McDonnell Foundation (FE).